\documentclass[letterpaper]{article}
\usepackage{aaai20}
\usepackage{times}
\usepackage{helvet}
\usepackage{courier}
\usepackage[hyphens]{url}
\usepackage{graphicx}
\usepackage{graphicx}
\usepackage{amsmath}
\usepackage{amssymb}
\usepackage{mathrsfs}
\usepackage{bm}
\usepackage{physics}
\usepackage{algorithm}
\usepackage{algpseudocode}
\usepackage{xcolor}

\title{Robust Multi-Output Learning with Highly Incomplete Data via Restricted Boltzmann Machines}
\author{Giancarlo Fissore, Aur\'elien Decelle,
Cyril Furtlehner \\
TAU-LRI-CNRS-INRIA \\
Laboratoire de Recherche en Informatique, \\
B\^at 660 Universit\'e Paris-Sud, \\
91405 Orsay Cedex France \\
\And Yufei Han \\
Symmantec, \\
Sophia Antipolis
}

\newcommand{\V}{{\cal V}}
\newcommand{\Hi}{{\cal H}}
\newcommand{\Ob}{{\cal O}}
\newcommand{\M}{{\cal M}}

\newcommand{\red}[1]{\textcolor{red}{#1}}
\DeclareMathOperator*{\argmax}{\text{argmax}}

\begin{document}


\maketitle              

\begin{abstract}
In a standard multi-output classification scenario, both features and labels of training data are partially observed. This challenging issue is widely witnessed due to sensor or database failures, crowd-sourcing and noisy communication channels in industrial data analytic services. Classic methods for handling multi-output classification with incomplete supervision information usually decompose the problem into an imputation stage that reconstructs the missing training information, and a learning stage that builds a classifier based on the imputed training set. These methods fail to fully leverage the dependencies between features and labels.
In order to take full advantage of these dependencies we consider a purely probabilistic setting in which the features imputation and multi-label classification problems are jointly solved. Indeed, we show that a simple Restricted Boltzmann Machine can be trained with an adapted algorithm based on mean-field equations to efficiently solve problems of inductive and transductive learning in which both features and labels are missing at random. The effectiveness of the approach is demonstrated empirically on various datasets, with particular focus on a real-world Internet-of-Things security dataset.

\end{abstract}

\section{Introduction}
Modern machine learning techniques usually require large sets of fully observed and well labelled data for training, which is usually unrealistic in real-world applications. Co-occurrence of missing features and partially observed label information is a widely witnessed issue in industrial data analytic business. On one hand, training features are not complete or heavily corrupted due to unpredictable sensor / end-device failure, collapse of database and noisy communication channels. On the other hand, human and automatic annotators can not recognize the incomplete feature profiles. It is safe for them to leave the incomplete training instances unlabelled and provide labels to the fully observed data instances with noise-free feature profiles. Consequently, it is highly demanding for the machine learning system to tolerate co-occurrence of missing features and partially observed labels for robust learning in practical use. 


We study both \textbf{transductive} and \textbf{inductive} multi-output classification. Multi-output classification, including \textbf{multi-class} and \textbf{multi-label} learning tasks, outputs class labels with higher dimensional representation. They thus define a more sophisticated learning scenario compared to binary classification. The former associates an input instance to one class of a finitely defined class set, while the latter allows one instance to be associated to multiple labels simultaneously. In the transductive case, the learning objective is to infer the missing features and labels based on the observed ones (no separate test set is required). In the inductive case, the model is trained over a set of \textbf{incomplete training instances} and the classification is performed by inferring the labels associated to \textbf{incomplete test instances}. Though the problem of multi-output learning with semi-supervised labels has been discussed in previous works \cite{YuYinSun:2010,Bucak:2011}, they all assume features of both training and test instances to be fully observed and free from noise. 
In our work, we consider that both features and labels are missing completely at random, which means that the mask of missing information is assumed to be statistically independent on the data distribution.  

State-of-the-art accuracies for classification with incomplete or noise-corrupted features and partially observed labels are given by \textbf{CLE} \cite{Han:2018}, \textbf{NoisyIMC} \cite{Chiang:2015} and \textbf{MC-1} \cite{Cabral:2015}. In particular, \textbf{CLE} and \textbf{NoisyIMC} are able to conduct both transductive and inductive learning for multi-label classification, while \textbf{MC-1} is a transductive-only method. It is worth noting that \textbf{CLE} has been originally proposed to handle the issue of weak labeling in multi-label learning, where only positive labels are observed. \textbf{NoisyIMC} and \textbf{MC-1} can work in the semi-supervised learning scenario. However, none of the three methods can handle all the challenges co-currently raised in our work. First, all of these assume the testing instances to have fully observed feature profiles. They don't consider coping with incomplete testing instances by design. Second, all of them are designed specifically for multi-label learning and adapting them to multi-class classification is not straightforward.

More recently, methods based on Deep Latent Variable Models (DLVM) have been proposed to deal with missing data. In \cite{miwae}, the Variational Autoencoder \cite{vae} has been adapted to be trained with missing data and a sampling algorithm for data imputation is proposed. Other approaches based on Generative Adversarial Networks (GAN) by \cite{gan} are proposed in \cite{gain} and \cite{misgan}. 
Impressive results on image datasets are displayed for these models, at the price of a rather high model complexity and the need for a large training set. In addition these works are focused on features reconstruction, and additional specifications and fine-tuning are required to be able to take partially observed labels into account. The models specifications are quite involved and any new specificity of the dataset may increase both the cost and the difficulty in training (especially for the approaches based on GANs).


In this paper we choose to address this problem in a more economical and robust manner. We consider the old and simple architecture of the Restricted Boltzmann Machine and adapt it to the multi-output learning context (RBM-MO) with missing data. 
The \textbf{RBM-MO} method serves as a generative model which collaboratively learns the marginal distribution of features and label assignments of input data instances, despite the incomplete observations. 
Building on the ideas expressed in \cite{Kappen94,em_supervised} we adapt the approach to the more effective contrastive divergence training procedure \cite{Hinton_CD} and provide results on various real-world datasets. The advantage of the RBM-MO model is that of providing a robust and flexible method to deal with missing data, with little additional complexity with respect to the classic RBM. Indeed, the trained model can be naturally applied to both transductive and inductive scenarios, achieving superior multi-output classification performance then state-of-the-art baselines. Moreover, it works seamlessly with multi-class and multi-label tasks, providing a unified framework for multi-output learning. 

In Section \ref{sec:rbm_intro} we will present a brief overview of RBM. In Section \ref{sec:rbm_algo} and \ref{sec:meanfield} we will show how the RBM can be adapted to work with missing data and we will present an effective imputation procedure. In Section \ref{sec:experiments} we present the results of our experimental study. Here, in addition to presenting results on some commonly used public datasets, 
we will also introduce a dataset for security in the Internet-of-Things which serves as a typical example of multi-output classification with highly incomplete data \cite{hidden} and evaluate
to what extent the proposed RBM-MO method can improve the quality of security services. 

\section{Overview of Restricted Boltzmann Machines} \label{sec:rbm_intro}


An RBM is a Markov random field with pairwise interactions defined on a bipartite graph formed by two layers of non-interacting variables: the visible nodes represent instances of the input data while the hidden nodes provide a latent representation of the data instances. $\V$ and $\Hi$ will denote respectively the sets of visible and hidden variables. In our setting, the visible variables will further split into two subsets $\V_f$ and  $\V_\ell$ corresponding respectively to features and labels, such that  $\V = \V_f+\V_\ell$.
The visible variables form an explicit representation of the data and are noted $\bm{v} = \{v_i, i\in\V\}$. The hidden nodes $\bm{h} = \{h_j, j\in\Hi\}$ serve to approximate the underlying dependencies among the visible units. 

In this paper, we will work with binary hidden nodes $h_j \in \{0,1\}$. The variables corresponding to the visible features will be either real with a Gaussian prior or binary, depending on the data to model, and labels variables will always be binary $\left( v_i \in \{0,1\} \right)$. 
The energy function of the RBM is defined as
\begin{equation}
E(\bm{v},\bm{h}) = - \sum_{i\in\V,j\in\Hi} v_i w_{ij} h_j - \sum_{i\in\V} a_i v_i - \sum_{j\in\Hi} b_j h_j
\label{eq:ef}
\end{equation}
where \(a_i\) and \(b_j\) are biases acting respectively on the visible and hidden units and $w_{ij}$ is the weight matrix that couples visible and hidden nodes. The joint probability distribution over the nodes is then given by the Boltzmann measure
\begin{equation}
\label{eq:rbm}
P(\bm{v},\bm{h}) = \frac{e^{-E(\bm{v},\bm{h})}}{Z} p_{\rm prior}(\bm{v})
\end{equation}
where $p_{\rm prior}$ is in product form and encodes the nature of each visible variable, either with a Gaussian prior $p_{\rm prior}= {\mathcal N}(0,\sigma_v^2)$ or a binary prior $p_{\rm prior}(v)= \delta(s^2-s)$.
$ \textstyle Z = \sum_{\bm{v},\bm{h}} p_{\rm prior}(\bm{v})e^{-E(\bm{v},\bm{h})}$ is the partition function. The classical training method consists in maximizing the marginal likelihood over the visible nodes $P(\bm{v}) = \sum_{\bm{h}} P(\bm{v},\bm{h})$ by tuning the RBM parameters $\theta = \{w_{ij},a_i,b_j\}$ via gradient ascent of the log likelihood $\mathcal{L}(\bm{v};\theta)$.  
The update rules for the weights (and similarly for the fields) are 
\begin{equation}
\Delta w_{ij} = \eta \frac{\partial\mathcal{L}(\bm{v};\theta)}{\partial w_{ij}}\label{eq:update}
\end{equation}
with the parameter $\eta$ corresponding to the learning rate. 

The tractability of the method relies heavily on the fact that the conditional probabilities $P(\bm{v}\vert\bm{h})$ and $P(\bm{h}\vert \bm{v})$ are given in closed forms.
In our case these read:
\begin{align}
P(\bm{v}\vert\bm{h}) = \prod_{i\in\V_f} &\frac{e^{\sum_{j\in\Hi}v_i w_{ij}h_j+a_iv_i} p_{\rm prior}(v_i)}{\sum_{v_i} e^{\sum_{j\in\Hi}v_i w_{ij}h_j+a_iv_i} p_{\rm prior}(v_i)}\nonumber\\[0.2cm] 
&\times\prod_{i\in\V_\ell} \sigma\Big(\sum_{j\in\Hi}v_i w_{ij}h_j+a_iv_i\Big),\label{eq:pvh}\\[0.2cm]
P(\bm{h}\vert\bm{v}) = \prod_{j\in\Hi} &\sigma\Big(\sum_{i\in\V}v_i w_{ij} + b_j\Big),\label{eq:phv}
\end{align}
where $\sigma(x) = 1/(1+e^{-x})$ is the logistic function.
The gradient of the likelihood w.r.t. the weights (and similarly w.r.t. the fields $a_i$ and $b_j$) is given by 
\begin{equation}
    \frac{\partial\mathcal{L}(\bm{v};\theta)}{\partial w_{ij}} = \langle v_i h_j p(h_j |\bm{v}) \rangle_{\rm data} - \langle v_i h_j \rangle_{\rm RBM}\label{eq:vanilladL}
\end{equation}
where the brackets $\langle\rangle_{\rm data}$ and $\langle\rangle_{\rm RBM}$ respectively indicate the average over the data and over the distribution \eqref{eq:rbm}. The positive term is directly linked to the data and can be estimated exactly with (\ref{eq:phv}), while the negative term is intractable.
Many strategies are used to compute this last term: the \emph{contrastive divergence} (CD) approach~\cite{Hinton_CD} consists in estimating the term over a finite number of Gibbs sampling steps, starting from a data point and making alternate use of (\ref{eq:pvh}) and (\ref{eq:phv}); in its \emph{persistent} version (PCD)~\cite{PCD} the chain is maintained over subsequent mini-batches; using mean-field approximation~\cite{TAP_train} the term is computed by means of a low-couplings expansion.    

\section{Learning RBM with incomplete data}\label{sec:rbm_algo}
The RBM is a generative model able to learn the joint distribution of some empirical data given as input. As such, it is intrinsically able to encode the relevant statistical properties found in the training data instances that relate features and labels, and this makes the RBM particularly suitable to be used in the multi-output setting in the presence of incomplete observations. In this sense, the most natural way to deal with incomplete observations is to marginalize over the missing variables; in this section we show how the contrastive divergence algorithm can be adapted to compute such marginals.

Given a partially-observed instance $\bm{v}$, we have a new partition of the visible space $\V = \Ob+\M$, where $\Ob$ is a subset of observed values of $\bm{v}$ that can correspond both to features and labels. $\bm{v}_o = \{v_i, i\in \Ob\}$ and $\bm{v}_m = \{v_i, i\in \M\}$ denote respectively the observed and missing values of $\bm{v}$. Let's define the following quantity 
\begin{align*}
Z_\Ob[\theta] &= \int \prod_{i\in \M}p_{\rm prior}(v_i)dv_i\\[0.2cm] 
&\times e^{\sum_{k\in\V} a_k v_k} \prod_{j\in\Hi} \left(1+\exp\Big( \sum_{k\in\V} w_{kj} v_k + b_j \Big)\right)
\end{align*}
corresponding to the marginalization of the numerator of the probability distribution over the visible variables, $\theta$ representing the parameters of the model. The probability over the observed variables \(\bm{v}_o\) is given by 
\[
P(\bm{v}_o) = \frac{Z_\Ob[\theta]}{Z_\emptyset[\theta]}.
\]
Taking the log-likelihood and then computing the gradient with respect to the weight matrix element $w_{ij}$ (also similarly for the fields $a_i$ and $b_j$), we obtain two different expressions for $i\in\Ob$ and $i\in\M$.
\begin{align}
\frac{\partial \log Z_\Ob[\theta]}{\partial w_{ij}} &= v_i \sum_{h_j} h_j p(h_j\vert\bm{v}_o),\qquad i\in\Ob\label{eq:pos_termO}\\[0.2cm] 
\frac{\partial \log Z_\Ob[\theta]}{\partial w_{ij}} &=  \sum_{h_j}\int dv_i v_i h_j p(v_i,h_j\vert\bm{v}_o).\qquad i\in\M\label{eq:pos_termM}
\end{align}
The gradient of the LL over the weights~(\ref{eq:vanilladL}) now reads
\begin{align}\label{eq:lossy_up}
\frac{\partial {\mathcal L}(\bm{v};\theta)}{\partial w_{ij}} &= \Big\langle I_o(i)  v_i \sum_{h_j} h_j p(h_j\vert\bm{v}_o) \Big\rangle_{\rm data} \nonumber \\[0.2cm]
&+\Big\langle\bigl(1-I_o(i)\bigr)\sum_{h_j}\int dv_i v_i h_j p(h_j\vert\bm{v}_o) \Big\rangle_{\rm data} \nonumber \\[0.2cm]
&-\langle v_ih_j\rangle_{\rm RBM}
\end{align}
where \(I_o\) is the indicator function of the samples dependent set $\Ob$.
The observed variables \(v_i, i \in \Ob\) are pinned to the values given by the training samples. In terms of our model, the pinned variables play the role of an additional bias over the hidden variables of a RBM where the ensemble of visible variables is reduced to the missing ones. This results in the following shift to the bias of the hidden nodes:
\begin{equation}
b_j \rightarrow b_j + \sum_{i \in \Ob} w_{ij} v_i
\end{equation}
With respect to the non-lossy case where $p(h_j\vert \bm{v})$ is given in closed form, here we need to sum over the missing variables in order to estimate $p(h_j\vert \bm{v}_o)$.
This means that also the positive term of the gradient \eqref{eq:lossy_up} is now intractable and we need to approximate it. For CD training, we can simply perform Gibbs sampling over the missing variables (keeping fixed the observed variables). The training algorithm then becomes:

\begin{algorithm}[H]
\caption{Lossy-CDk (RBM training with Incomplete data)}\label{alg:lcdk}
\begin{algorithmic}[1]
\State \textbf{Data:} a training set of N data vectors
\State Randomly initialize the weight matrix \textbf{W}
\For{t = 0 to T (\# of epochs)}
    \State Divide the training set in \(m\) minibatches
    \ForAll{minibatches \(m\)}

    \textbf{Positive term:}
    \State pin variables \(v_i, i \in \Ob\) to their correct value
    \State initialize \(v_i, i \in \M\) randomly
    \State sample \(\bm{h}, \bm{v}_m\) using \(p(\bm{v}_m\mid \bm{h})\) and \(p( \bm{h} \mid \bm{v})\) for \textit{k} steps
    \State compute the positive terms in \eqref{eq:pos_termO} and \eqref{eq:pos_termM}

    \textbf{Negative term:}
	\State initialize \(\bm{v}\) randomly
	\State iterate eq. \eqref{eq:pvh}, \eqref{eq:phv} (\textit{k} steps) to compute \(\langle v_i h_j \rangle_{model}\)

	\textbf{Full update:} \State update \(\mathbf{W}\) with equations \eqref{eq:update} and \eqref{eq:lossy_up}
  \EndFor
\EndFor
\end{algorithmic}
\end{algorithm}

We note that the extra computational burden of Lossy-CD with respect to standard CD is due only to the extra Gibbs sampling steps in the positive term. Given that the observed variables strongly bias the sampling procedure speeding up convergence, only few sampling steps are needed to compute this term. Indeed, in our experiments we observed that a single sampling step (Lossy-CD1) is enough, making the additional complexity minimal. Finally, we note that the same method can be applied to PCD and mean-field training procedures. In the first case, it is sufficient to keep track of an additional persistent chain, which requires little extra memory and no extra computational complexity. In the second case, we only need to substitute Gibbs sampling with iterative mean-field equations.

\section{Mean-field based imputation with RBM}\label{sec:meanfield}
As a generative model, the trained RBM can be used to sample new data. For imputation of missing features and labels we just need to use the observed portions of our data to bias the sampling procedure in the same way as for the computation of the positive term in Alg. 1. Namely, we estimate \(p(\bm{v_m}\vert \bm{v_o})\) by pinning the observed variables and iterating CD/PCD or mean-field to approximate the equilibrium values of the missing variables. In case of a high percentage of missing observations, however, we might expect the observed variables to be correlated to many different equilibrium configurations, such that the sampling could be biased towards the wrong sample. This effect is present also during training, but in that case it is mitigated by the average over the minibatch while for imputation over a single data instance it becomes relevant. To overcome this problem, we simply average over multiple imputations for each incomplete data instance. Here we highlight the fact that the equilibrium configurations of the RBM are weighted according to the empirical data distribution; as a consequence, the bias toward an incorrect sample is easily discarded and generally a small number of different imputations need to be averaged to obtain the correct result. 
To further reduce the number of imputations, we employ mean-field. Indeed, a good equilibrium configuration could require more CD/PCD samples (simply due to the sampling noise) while we expect a non-extensive number of fixed-points related to meaningful (equilibrium) configurations to be present \cite{DeFiFu2018} and these are directly obtained by iteration.  

More in details, let $\{p_i,i\in\V_\ell \}$ and $\{q_j,j\in \Hi\}$ be the marginal probabilities respectively of visible labels and hidden variables to be activated and $\{m_i,i\in\V_f \}$ the marginal expectation of the 
visible features variables. Mean-field equations at lowest order (${\cal O}(1/N)$, \(N\) being the size of the system) express self-consistent relations among these quantities
\begin{align}
    m_i &= \Big(\sum_{j\in\Hi}w_{ij}q_j+a_i\Big)\sigma_v^2,\qquad\forall i\in\V_f\backslash\Ob\label{eq:mi} \\[0.2cm]
    p_i &= \sigma\Big(\sum_{j\in\Hi}w_{ij}q_j+a_i\Big),\qquad\forall i\in\V_\ell\backslash\Ob\label{eq:pi}\\[0.2cm]
    q_j &= \sigma\Big(\sum_{i\in\V_f}w_{ij}m_i+\sum_{i\in\V_\ell}w_{ij}p_i+b_j\Big)\label{eq:qj}
\end{align}
Higher order terms corresponding to TAP equations are discarded \cite{Mezard}.
These equations can be efficiently solved by iteration starting from random configurations until a fixed point is reached. Observed variables are simply introduced by pinning their corresponding probabilities ($0$ or $1$ for label variables) or their marginal expectation (for feature variables) to the observed values. In practice we run these fixed-point equations $N_f\sim 10$ times and the imputations are obtained by 
simple average
\begin{align*}
\hat m_i &= \frac{1}{N_f} \sum_{n=1}^{N_f} m_i^{(n)}\\[0.2cm]
p_i &= \frac{1}{N_f} \sum_{n=1}^{N_f} p_i^{(n)}.
\end{align*}
In the multi-label setting, the predictor is the indicator function $\hat p_i = (p_i>t)$ (\(t\) is learned, it is chosen to maximize the accuracy for known labels), while for class labels we have
\[
\hat p_i =
\begin{cases}
1,\qquad \text{for}\ i =\argmax_k (p_k)\\[0.2cm]
0,\qquad\text{otherwise}.
\end{cases}
\]
\begin{table*}[ht]
\small
\caption{Transductive test on \textit{MNIST} multi-class data set\label{tab:tranductive_mc}}
\centering
\begin{tabular}[c]{|c|c|c|c|c|c|c|c|c|c|}
  \hline
    Model & \multicolumn{3}{|c|}{RMSE} & \multicolumn{3}{|c|}{Averaged AUC} & \multicolumn{3}{|c|}{Accuracy} \\\hline
  $q_{mc}\%$ & 30\% & 50\% & 80\% & 30\% & 50\% & 80\% & 30\% & 50\% & 80\% \\\hline

  
  \textbf{RBM-MO}($q_{fea}\%=$50\%)
    & \red{\textbf{0.183}} & \red{\textbf{0.182}} & \red{\textbf{0.185}}
    & \red{\textbf{0.969}} & \red{\textbf{0.971}} & \red{\textbf{0.929}}
    & \red{\textbf{0.950}} & \red{\textbf{0.912}} & \red{\textbf{0.822}} \\
    CLE($q_{fea}\%=$50\%) & 0.195 & 0.195 & 0.195 & 0.686 & 0.718 & 0.742 & 0.256 & 0.232 & 0.282 \\
    NoisyIMC($q_{fea}\%=$50\%) & 0.209 & 0.210 & 0.210 & 0.621 & 0.578 & 0.552 & 0.225 & 0.232 & 0.192 \\
    MC-1($q_{fea}\%=$50\%) & 0.334 & 0.335 & 0.337 & 0.495 & 0.493 & 0.500 & 0.110 & 0.111 & 0.112 \\\hline
    
  \textbf{RBM-MO}($q_{fea}\%=$80\%)
    & \textbf{0.209} & \textbf{0.213} & \textbf{0.211}
    & \red{\textbf{0.938}} & \red{\textbf{0.932}} & \red{\textbf{0.906}}
    & \red{\textbf{0.920}} & \red{\textbf{0.852}} & \red{\textbf{0.733}} \\
    CLE($q_{fea}\%=$80\%) 
        & \red{0.206} & \red{0.208} & \red{0.206}
        & 0.673 & 0.678 & 0.625
        & 0.230 & 0.215 & 0.220 \\
    NoisyIMC($q_{fea}\%=$80\%) & 0.212 & 0.211 & 0.213 & 0.652 & 0.577 & 0.537 & 0.230 & 0.217 & 0.210\\
    MC-1($q_{fea}\%=$80\%) & 0.334 & 0.334 & 0.335 & 0.500 & 0.501 & 0.500 & 0.112 & 0.110 & 0.110\\
  
  \hline
\end{tabular}
\end{table*}

\begin{table*}
\small
\caption{Transductive test on \textit{Scene} multi-label data set\label{tab:trans_scene}}
\centering
\begin{tabular}[c]{|c|c|c|c|c|c|c|c|c|c|}
  \hline
    Model & \multicolumn{3}{|c|}{RMSE} & \multicolumn{3}{|c|}{Micro-AUC} & \multicolumn{3}{|c|}{Hamming-Accuracy} \\
    \hline
    $q_{ml}\%$ & 30\% & 50\% & 80\% & 30\% & 50\% & 80\% & 30\% & 50\% & 80\% \\\hline
  
  \textbf{RBM-MO}($q_{fea}\%=$50\%)
    & \textbf{0.131} & \textbf{0.137} & \red{\textbf{0.123}}
    & \red{\textbf{0.943}} & \red{\textbf{0.934}} & \textbf{0.888}
    & \red{\textbf{0.919}} & \red{\textbf{0.907}} & \textbf{0.873} \\
  CLE($q_{fea}\%=$50\%)
    & \red{0.130} & \red{0.130} & 0.131
    & 0.905 & 0.893 & \red{0.898}
    & 0.885 & 0.871 & \red{0.878} \\
  NoisyIMC($q_{fea}\%=$50\%) & 0.132 & 0.133 & 0.133 & 0.865 & 0.863 & 0.858 & 0.845 & 0.841 & 0.848 \\
  MC-1($q_{fea}\%=$50\%) & 0.258 & 0.255 & 0.267 & 0.522 & 0.528 & 0.527 & 0.826 & 0.817 & 0.824 \\\hline
  
  \textbf{RBM-MO}($q_{fea}\%=$80\%)
    & \textbf{0.160} & \textbf{0.158} & \textbf{0.158}
    & \textbf{0.875} & \textbf{0.867} & \textbf{0.826}
    & \textbf{0.856} & \textbf{0.858} & \textbf{0.832} \\
  CLE($q_{fea}\%=$80\%)
    & \red{0.129} & \red{0.129} & \red{0.128}
    & \red{0.913} & \red{0.897} & \red{0.899}
    & \red{0.889} & \red{0.875} & \red{0.876} \\
  NoisyIMC($q_{fea}\%=$80\%) & 0.133 & 0.134 & 0.134 & 0.853 & 0.857 & 0.849 & 0.839 & 0.835 & 0.826 \\
  
  \hline
\end{tabular}
\end{table*}

\begin{table*}[h]
\small
\caption{Transductive test on \textit{EventCat} multi-label data set\label{tab:eventcat_transductive}}
\centering
\begin{tabular}[c]{|c|c|c|c|c|c|c|c|c|c|}
  \hline
    Model & \multicolumn{3}{|c|}{RMSE} & \multicolumn{3}{|c|}{Micro-AUC} & \multicolumn{3}{|c|}{Hamming-Accuracy} \\
    \hline
    $q_{ml}\%$ & 30\% & 50\% & 80\% & 30\% & 50\% & 80\% & 30\% & 50\% & 80\% \\\hline
  
  \textbf{RBM-MO}($q_{fea}\%=$50\%) 
    & \textbf{2.316} & \textbf{2.427} & \textbf{2.463}
    & \textbf{0.842} & \textcolor{red}{\textbf{0.852}} & \textbf{0.794}
    & \textcolor{red}{\textbf{0.972}} & \textcolor{red}{\textbf{0.956}} & \textcolor{red}{\textbf{0.912}} \\
  CLE($q_{fea}\%=$50\%) & 9.543 & 8.434 & 10.902 & \textcolor{red}{0.872} & 0.837 & 0.739 & 0.900 & 0.870 & 0.800 \\
  NoisyIMC($q_{fea}\%=$50\%) & 8.443 & 8.792 & 9.850 & 0.835 & 0.817 & \textcolor{red}{0.799} & 0.820 & 0.801 & 0.738 \\
  MC-1($q_{fea}\%=$50\%) & \textcolor{red}{2.220} & \textcolor{red}{2.180} & \textcolor{red}{2.128} & 0.590 & 0.607 & 0.573 & 0.697 & 0.700 & 0.693 \\\hline
  
  \textbf{RBM-MO}($q_{fea}\%=$80\%)
    & \textbf{3.235} & \textbf{3.128} & \textbf{3.238}
    & \textcolor{red}{\textbf{0.843}} & \textbf{0.782} & \textbf{0.746}
    & \textcolor{red}{\textbf{0.972}} & \textcolor{red}{\textbf{0.945}} & \textcolor{red}{\textbf{0.910}} \\
  CLE($q_{fea}\%=$80\%) & 15.265 & 15.989 & 14.559 & 0.832 & \textcolor{red}{0.784} & \textcolor{red}{0.785} & 0.885 & 0.757 & 0.729 \\
  NoisyIMC($q_{fea}\%=$80\%) & 14.155 & 14.289 & 14.395 & 0.852 & 0.822 & 0.803 & 0.895 & 0.807 & 0.735 \\
  MC-1($q_{fea}\%=$80\%) & \textcolor{red}{2.776} & \textcolor{red}{2.423} & \textcolor{red}{2.476} & 0.536 & 0.540 & 0.536 & 0.691 & 0.688 & 0.687 \\
  \hline
\end{tabular}
\end{table*}

\begin{table*}
\small
\centering
\caption{Inductive test on \textit{Pendigits} multi-class dataset\label{tab:inductive_pen}}
\begin{tabular}[c]{|c|c|c|c|c|c|c|}
  \hline
    Model & \multicolumn{3}{|c|}{Averaged AUC} & \multicolumn{3}{|c|}{Accuracy} \\
    \hline
    $q_{mc}\%$ & 30\% & 50\% & 80\% & 30\% & 50\% & 80\% \\\hline
  
  \textbf{RBM-MO}($q_{fea}\%=$50\%)
    & \red{\textbf{0.887}} & \red{\textbf{0.914}} & \red{\textbf{0.910}}
    & \red{\textbf{0.533}} & \red{\textbf{0.673}} & \red{\textbf{0.660}}\\
  CLE($q_{fea}\%=$50\%) & 
  0.785 & 0.791 & 0.791 & 0.297 & 0.256 & 0.268 \\
  NoisyIMC($q_{fea}\%=$50\%) &
  0.780 & 0.771 & 0.781 & 0.302 & 0.272 & 0.265\\\hline
  
  \textbf{RBM-MO}($q_{fea}\%=$80\%)
    & \red{\textbf{0.891}} & \red{\textbf{0.909}} & \red{\textbf{0.889}}
    & \red{\textbf{0.562}} & \red{\textbf{0.682}} & \red{\textbf{0.647}} \\
  CLE($q_{fea}\%=$80\%) &
  0.768 & 0.664 & 0.622 & 0.271 & 0.200 & 0.176 \\
  NoisyIMC($q_{fea}\%=$80\%) &
  0.748 & 0.687 & 0.615 & 0.264 & 0.220 & 0.178\\
  
  \hline
\end{tabular}
\end{table*}

\begin{table*}
\small
\centering
\caption{Inductive test on \textit{EventCat} multi-label data set\label{tab:eventcat_inductive}}
\begin{tabular}[c]{|c|c|c|c|c|c|c|}
  \hline
    Model & \multicolumn{3}{|c|}{Micro-AUC} & \multicolumn{3}{|c|}{Hamming-Accuracy} \\
    \hline
   $q_{ml}\%$ & 30\% & 50\% & 80\% & 30\% & 50\% & 80\% \\\hline

  
  \textbf{RBM-MO}($q_{fea}\%=$50\%)
    & \textcolor{red}{\textbf{0.839}} & \textcolor{red}{\textbf{0.826}} & \textcolor{red}{\textbf{0.793}}
    & \textcolor{red}{\textbf{0.970}} & \textcolor{red}{\textbf{0.970}} & \textcolor{red}{\textbf{0.965}} \\
  CLE($q_{fea}\%=$50\%) & 0.705 & 0.707 & 0.706 & 0.700 & 0.724 & 0.719 \\
  NoisyIMC($q_{fea}\%=$50\%) & 0.704 & 0.702 & 0.700 & 0.710 & 0.717 & 0.718 \\\hline
  
  \textbf{RBM-MO}($q_{fea}\%=$80\%)
    & \textcolor{red}{\textbf{0.759}} & \textcolor{red}{\textbf{0.791}} & \textcolor{red}{\textbf{0.766}}
    & \textcolor{red}{\textbf{0.964}} & \textcolor{red}{\textbf{0.964}} & \textcolor{red}{\textbf{0.967}} \\
  CLE($q_{fea}\%=$80\%) & 0.693 & 0.688 & 0.694 & 0.718 & 0.706 & 0.718 \\
  NoisyIMC($q_{fea}\%=$80\%) & 0.689 & 0.688 & 0.685 & 0.705 & 0.704 & 0.704 \\
  
  \hline
\end{tabular}
\end{table*}

\section{Experimental Study}\label{sec:experiments}
\subsection{Experimental configuration}

To evaluate the efficiency of RBM-MO we compare its performance against \textbf{CLE}, \textbf{NoisyIMC} and \textbf{MC-1}, which provide state-of-the-art baselines.

For the transductive experiments we randomly hide features and labels of the whole dataset to generate incomplete data for training, and we compute appropriate scores for the reconstruction of missing features and labels. In the inductive test, instead, we split the whole dataset into non-overlapping training and testing sets. Concerning the training set the same protocol is used as in the transductive test. For the test set the difference is that now all labels are hidden. Once the classifier is trained, it is applied on the test set to predict the labels.
We still randomly hide the entries of test features vectors, so as to form an \textbf{incomplete testing set}.
Finally, in the splitting we use 70\% of the data instances for training and the remaining 30\% for testing.

We denote by $q_{fea}$, $q_{ml}$ and $q_{mc}$ the percentage of masked features, labels and classes labels respectively. Note that a masked  class label means that all binary variables attached to the classes of a given label are masked together.    
These rates of masking are kept identical in the learning and test sets.

In the transductive test, we compute the Root Mean Squared Error (RMSE) to measure the reconstruction accuracy with respect to the missing feature values, defined w.r.t. the dataset ${\cal S}$ as: 
\begin{equation}
    RMSE = \sqrt{\frac{1}{\vert\cal S\vert\vert\bar{\cal O}\vert}\sum_{s,i \in {\cal S}\times\bar{\cal O}} |X_{s,i} - \hat{X}_{s,i}|^{2}} \nonumber
\end{equation}
where $X$ and $\hat{X}$ are the incomplete and reconstructed feature matrices. $\bar {\cal O}$ denotes the set of unobserved features. Furthermore, for the reconstructed labels we calculate \textbf{Micro-AUC} scores and \textbf{Hamming-accuracy} \cite{Madjarov2012} in the multi-label scenario, and \textbf{Averaged AUC} plus \textbf{Accuracy} \cite{Li:2015} in the multi-class case. In the tables, we define \textbf{Hamming-accuracy} as \textbf{1-Hamming loss} to keep a consistent variation tendency with the AUC scores. In the inductive test we only compute the scores on the reconstructed labels, since reconstructing missing features is not the goal of inductive classification.

We run the test as described $10$ times with different realizations of the missing features and labels. Average and standard deviation of the computed scores are recorded to compare the overall performances. In the tables, we use red fonts to denote the best reconstruction and classification performances among all the algorithms involved in the empirical study. The bold black font is used to highlight the performance of the proposed \textbf{RBM-MO} method.

For the baselines, we used grid search to choose the optimal parameter combination following the suggested ranges of parameters as in \cite{Han:2018}. 

The RBM-MO is trained following the guidelines in \cite{Hinton:2010}. We always use binary variables for the hidden layer, while in the visible layer we use binary variables for MNIST and Gaussian variables for the other datasets.
In all the simulations, we fix the number of hidden nodes to 100. The learning rate \(\eta\) is fixed to 0.001 and the size of the mini-batches to 10. During training the number of Gibbs steps is set to \(k = 1\) while for imputation we iterate the mean-field equations 10 times. As a stopping condition, we considered the degradation of the transductive AUC scores with a look-ahead of 500 epochs\footnote{Full code with instructions for reproducibility and extensions is released: \(\langle\) link hidden for blind review \(\rangle\)}.

\subsection{Summary of datasets}
We consider 4 publicly available datasets emerging from diverse disciplines, such as image processing and biology. These datasets cover both multi-label and multi-class learning tasks, and they are popularly used as benchmark datasets in multi-output learning research.

\begin{table}[]
\centering
\small
\caption{Summary of 4 public multi-label and multi-class data sets.}
\label{tab:datasets}
\resizebox{\linewidth}{!} {
\begin{tabular}{|c|c|c|c|c|}
\hline
\textbf{Dataset} & \textbf{No. of Instances} & \textbf{No. of Features} & \textbf{No.of Labels} & \textbf{No. of Classes}\\ \hline
   
   Yeast &        2,417          &      103     & 14 & -   \\ \hline 
   Scene    &      2,407            &   294  &   6 & -     \\ \hline 
   Pendigits  &  10992 & 16 & - & 10 \\\hline 
     MNIST   &        70,000          &       784 & - & 10           \\ \hline
     EventCat &        5,93          &      72     & 6 & -   \\ \hline
\end{tabular}
}
\vspace{-0.5cm}
\end{table}

In addition, we consider the challenging scenario of abnormality detection on IoT devices.
The relevant dataset consists in security telemetry data collected from various network
appliances (e.g. smart watches, smartphones, driving assistance systems...), each reporting a features vector
whose entries indicate the occurring frequency of a specific type of alert (e.g. downloading suspicious files, login failures, unfixed vulnerabilities...).
Multiple labels are assigned to each device in the collected dataset, corresponding
to a variety of categories of security threats. The learning problem is cast as a multi-label classification as the same telemetry features
can be related to multiple threats (e.g. scanning
activity and data breaching can occur simultaneously)
and the collected records for training and testing are usually highly incomplete due to various causes. For example, customers can set up their privacy control policies to limit the coverage of the telemetry data shared with security vendors, or deployed sensors can unexpectedly fail. Simultaneously, it is safe for human analysts to only label the most typical and clearly defined cyber menaces, as mislabeling can reduce usability of the provided protection service by causing false alarm and/or false negative. 
In the following, we will test our method on a complete realization of such a dataset that we refer to as \textit{EventCat}.

Some details about the datasets are reported in Table.\ref{tab:datasets}.
 

\subsection{Qualitative results on MNIST}

\begin{figure}[ht]
  \begin{center}
     
    $\begin{array}{ccccc}
      \includegraphics[width=.15\linewidth]{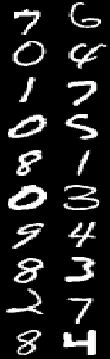} &
      \includegraphics[width=.15\linewidth]{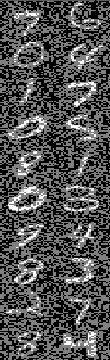} &
      \includegraphics[width=.15\linewidth]{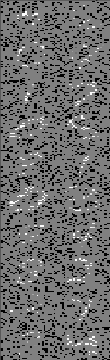} &
      \includegraphics[width=.15\linewidth]{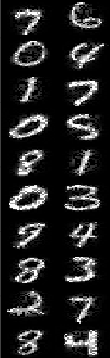} &
      \includegraphics[width=.15\linewidth]{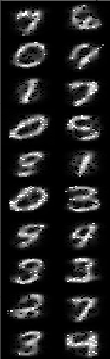}
    \end{array}$
  \end{center}
  \caption{Features reconstruction by RBM-MO trained over an incomplete dataset with 50\% missing-at-random features, whose classification accuracy has been measured to be around 91\%. The first block shows some complete testing instances. The second and third block show the same testing instances after hiding respectively 50\% and 80\% of the pixels. The last two columns show the results of the mean-field imputations over the incomplete testing instances.}
  \label{fig:reconstructions}
\end{figure}

A qualitative evaluation of the performance of the RBM-MO model is given by looking at features reconstruction for the MNIST dataset, as reported in Fig. \ref{fig:reconstructions}. The model at hand has been trained over a dataset in which 50\% of the features were missing. To assess the robustness of the method, we computed the reconstructions in the highly challenging case in which 80\% of the features were missing. Apart from some smoothing due to the employment of mean-field imputations, the reconstructed samples look reasonably realistic. In general, from the qualitative point of you the results are comparable to those obtained with more complex and expensive DLVMs like MIWAE and MisGAN \cite{miwae,misgan}.

\subsection{Empirical study of transductive learning on public datasets}

The transductive results for MNIST (multi-class) and Scene (multi-label) datasets are reported in tables \ref{tab:tranductive_mc} and \ref{tab:trans_scene}, while for the other datasets the full results are reported in the supplementary material. Globally the results show better performances of \textbf{RBM-MO} compared to the baseline methods \textbf{CLE}, \textbf{NoisyIMC} and \textbf{MC-1}. 
Going into the details, we first observe that 
\textbf{RBM-MO} is by a large margin more efficient than all of the baselines for the inference of class labels (see table \ref{tab:tranductive_mc} and the Pendigits table in the supplementary material), probably because it is able to encode more complex statistical properties.
Among the baseline methods, \textbf{CLE} performs the best. Explicitly enforcing a predictive constraint on the subspace representation of features and labels in the model, provides more robustness against missing information.  

On the multi-label problems, the situation is still in favour of  \textbf{RBM-MO} but with less margin (table \ref{tab:trans_scene} and Yeast table in the supplementary material), in particular at a larger percentage of missing features.

Now if we look at the reconstruction error on these datasets we observe that \textbf{RBM-MO} generally achieves a higher reconstruction accuracy than the other opponents, especially on the MNIST dataset. The results verify empirically the basic motivation of using a generative model, such as the RBM, in the raised challenging learning scenarios: \textbf{incomplete features and labels can provide complementary information to each other, so as to better recover the missing elements}. The variance of the results is omitted in the tables by lack of space. For \textbf{RBM-MO} the standard deviation of the derived RMSE, AUC and accuracy scores is not larger than $0.01$ over the different datasets. Although the RMSE scores reported by the baseline methods look comparable to the RBM-MO ones, and in certain cases they are better, they also come with a slightly higher variance, such that the RBM-MO seems to be more efficient and robust for features reconstruction.

\subsection{Empirical study of inductive learning on public datasets}



  

Except \textbf{MC-1}, all the other baseline methods are used for inductive learning. As in the transductive test, we show only the mean of the derived metrics in the tables. Nevertheless, we have similar variance ranges for the computed scores as reported in the transductive test. 
Clearly \textbf{RBM-MO} is much better adapted to this setting than the baseline methods both for multi-class (table \ref{tab:inductive_pen}) and multi-label learning (Yeast and Scene tables in supplementary material). With a few exceptions at large missing rates ($80\%$) the results of \textbf{RBM-MO} are distinctively better. 
The baseline inductive methods \textbf{CLE} and \textbf{NoisyIMC} are specifically designed for multi-label learning. In the multi-class scenario, unlabeled instances can't be treated properly. If a data instance is unlabeled, the whole corresponding row in the label matrix will be considered as missing. These structured missing patterns violate the assumption of random entry missing in semi-supervised learning, which thus leads to performance deterioration of the baseline methods. By comparison, \textbf{RBM-MO} can be adapted seamlessly to multi-class and multi-label learning, producing consistently good performances. 

%

\subsection{Transductive and inductive evaluation on \textit{EventCat} data}


Transductive results are reported in table \ref{tab:eventcat_transductive}. Compared to the three baseline approaches, \textbf{RBM-MO} shows an overall improvement of the combined features and labels reconstruction accuracy for all combinations of $\{q_{fea},q_{ml}\}$. When not the best one (at low missing percentage) \textbf{RBM-MO} reaches comparable features reconstruction errors to the best baseline. Meanwhile the classification performance is consistently better, with generally better AUC scores and up to $30\%$ higher hamming accuracy scores compared to baseline methods. Note that \textbf{MC-1} achieves the best features imputation results in the transductive test when $q_{fea} \geq {50\%}$. However \textbf{RBM-MO} can achieve a similar features imputation accuracy while producing significantly higher labels recovery precision. 

As seen in table \ref{tab:eventcat_inductive}, the results of inductive test of \textbf{RBM-MO} are equally encouraging. 
Even with highly incomplete training data, e.g. $q_{fea} = 0.5$ and $q_{ml} = 0.5$, \textbf{RBM-MO} produces the best predictions over partially observed testing data instances. In the extreme situation with $q_{fea}$ and $q_{ml}$ set to 0.8, the performances of all the methods are relatively low. Although the algorithms can not work on a stand-alone basis in this situation, their output can guide human analysts to effectively narrow down the hypotheses of correlation between event profiles and threat types. Despite the difficult learning scenario, \textbf{RBM-MO} still achieves the highest precision, which denotes its applicability for aiding human analysts to identify useful heuristic rules for detecting malicious events.



\section{Conclusion}\label{sec:conclusion}
Machine learning is witnessing a race to high complexity models eager for large data and computational power. 
In the context of multi-output classification in a challenging scenario - (i) learning with highly incomplete features and partially observed labels; ii) applying the learnt classifier with incomplete testing instances) - 
we advocate instead for simple probabilistic and interpretable models. After refining the learning of the RBM model, 
we give empirical evidences that it can be efficiently adapted to this context on a great variety of datasets. Experiments are conducted on both public databases and a real-world IoT security dataset, showing various sizes of training sets as well as features and labels vectors. Our approach consistently outperforms the state-of-the-art robust multi-class and multi-label learning approaches with imperfect training data, indicating good usability for practical applications.

\bibliography{rbm}
\bibliographystyle{aaai}

\end{document}